\documentclass[letterpaper, 10 pt, conference]{ieeeconf}  

\IEEEoverridecommandlockouts                              

\usepackage{algorithm}
\usepackage{hyperref}
\usepackage{xcolor}
\usepackage{algpseudocode}
\usepackage{graphics} 
\usepackage{epsfig} 
\usepackage{mathptmx} 
\usepackage{times} 
\usepackage{amsmath} 
\usepackage{amssymb}  

\title{\LARGE \bf
Advancing Robustness in Deep Reinforcement Learning with an Ensemble Defense Approach}

\author{Adithya Mohan$^{1}$, Dominik Rößle$^{1}$, Daniel Cremers$^{2}$ and Torsten Schön$^{1}$
\thanks{This work was supported by Hightech-Agenda Bayern}
\thanks{$^{1}$The Author is with AImotion Bavaria, Technische Hochschule Ingolstadt, Germany}%
\thanks{$^{2}$The author is with School of Computation, Information and Technology, TU Munich, Germany}%
}

\begin{document}

\maketitle
\thispagestyle{empty}
\pagestyle{empty}

\begin{abstract}
Recent advancements in Deep Reinforcement Learning (DRL) have demonstrated its applicability across various domains, including robotics, healthcare, energy optimization, and autonomous driving. However, a critical question remains: How robust are DRL models when exposed to adversarial attacks? While existing defense mechanisms such as adversarial training and distillation enhance the resilience of DRL models, there remains a significant research gap regarding the integration of multiple defenses in autonomous driving scenarios specifically. This paper addresses this gap by proposing a novel ensemble-based defense architecture to mitigate adversarial attacks in autonomous driving. Our evaluation demonstrates that the proposed architecture significantly enhances the robustness of DRL models. Compared to the baseline under FGSM attacks, our ensemble method improves the mean reward from \textbf{5.87} to \textbf{18.38} (over \textbf{213\%} increase) and reduces the mean collision rate from \textbf{0.50} to \textbf{0.09} (an \textbf{82\%} decrease) in the highway scenario and merge scenario, outperforming all standalone defense strategies.

\end{abstract}


\section{Introduction}
Reinforcement Learning (RL) has emerged as a pivotal methodology in developing autonomous driving systems, enabling vehicles to learn optimal decision-making strategies through interaction with their environment and feedback in the form of rewards \cite{Ilahi2022-bq}. When integrated with deep neural networks, Deep Reinforcement Learning (DRL) empowers agents to navigate complex, high-dimensional state and action spaces, facilitating significant advancements in autonomous driving technology. DRL has been instrumental in various autonomous driving tasks, including path planning, behavior modeling, traffic negotiation, and adaptive cruise control \cite{feng2021survey}.

Beyond autonomous driving, DRL has also shown success in other domains such as healthcare for personalized treatment strategies \cite{nguyen2019deep}, robotics for handling dynamic tasks \cite{nguyen2019deep}, energy systems for demand-response optimization \cite{lin2020review}, and the financial sector for portfolio management and fraud detection \cite{nguyen2019deep}.

Despite its transformative potential, DRL systems face significant challenges in real world applications, particularly in safety-critical domains. One major concern is their vulnerability to adversarial attacks strategically crafted inputs designed to exploit weaknesses in the model and manipulate the agent's behavior. For example, in autonomous driving, adversarial perturbations to sensory inputs such as camera or lidar data can cause an agent to veer off-road or ignore traffic signals \cite{huang2017adversarial}. In one study, small image perturbations caused an end-to-end DRL driving agent to make incorrect lane change decisions, potentially leading to collisions \cite{behzadan2017whatever}. Similarly, physical world attacks like applying stickers to stop signs can trick DRL based perception modules into misclassification \cite{evtimov2017robust}.

\begin{figure}
    \centering
    \includegraphics[width=260pt]{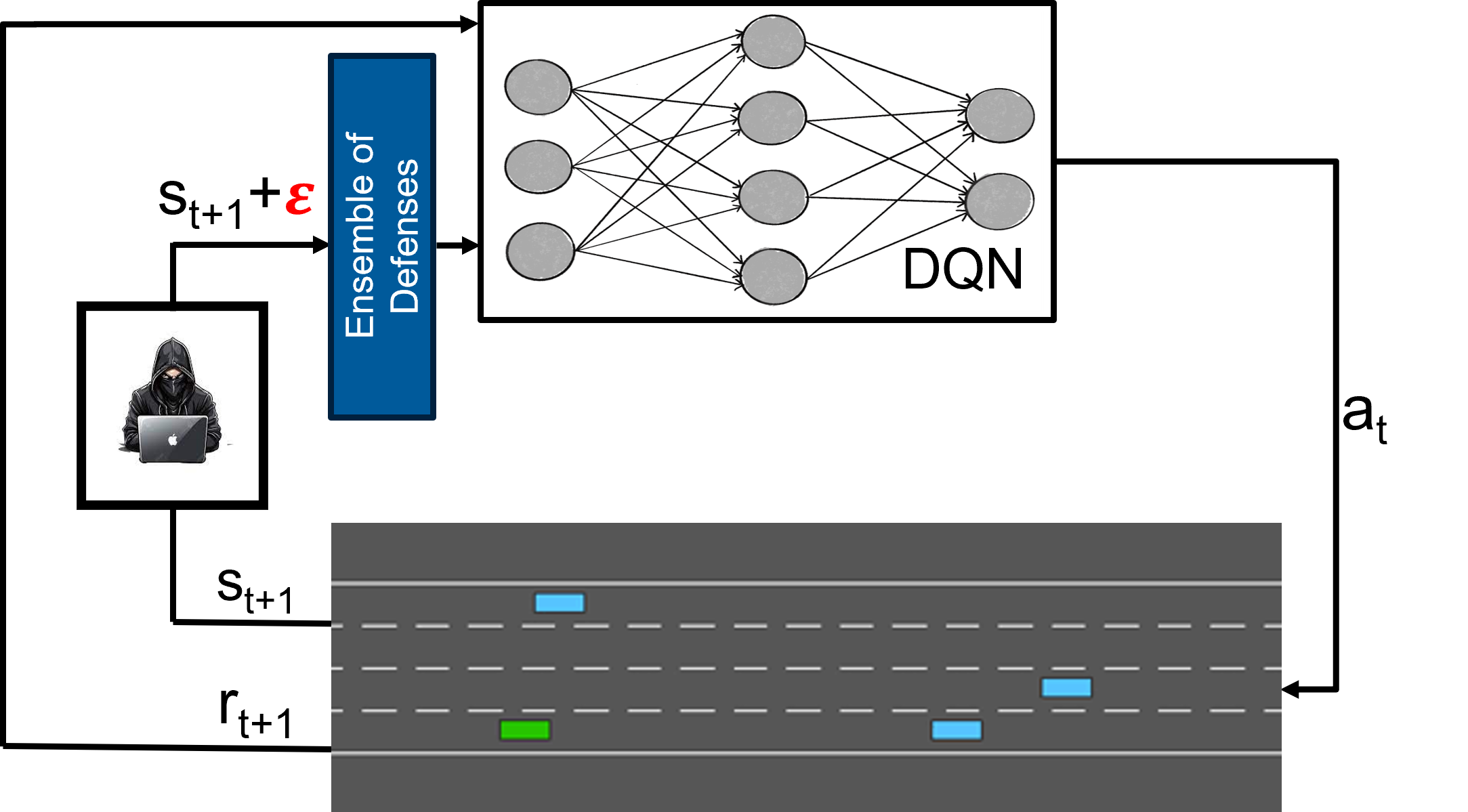}
    \caption{Overview of the proposed Ensemble Defense Framework for Deep Reinforcement Learning under adversarial attacks. During inference, the agent receives a perturbed observation ($\textcolor{red}{\epsilon}$) resulting from adversarial noise Fast Gradient Sign Method (FGSM). This perturbed state is simultaneously passed through three independent defense modules: (i) \textbf{Random Noise}, which introduces additional controlled noise to neutralize adversarial patterns, (ii) \textbf{Autoencoder}, which reconstructs the state using learned nominal representations, and (iii) \textbf{Principal Component Analysis (PCA)}, which projects the input onto a lower-dimensional subspace to suppress irrelevant noise. The outputs from these modules are aggregated via simple averaging to form a robust corrected observation, which is then used by the fixed DRL policy to select actions. The framework operates entirely at inference-time and requires no policy retraining, making it suitable for real-world deployment in safety-critical environments such as autonomous driving.}
    \label{fig:ensemble_defense}
\end{figure}

These vulnerabilities raise serious concerns about the reliability of DRL in real world deployment. In healthcare, for example, adversarial inputs could lead to incorrect treatment recommendations. In autonomous driving, they could result in erratic or unsafe driving behaviors, especially in dense traffic or urban environments. As DRL becomes increasingly integrated into such critical systems, ensuring their trustworthiness is essential to prevent catastrophic outcomes and promote public acceptance \cite{goodfellow2014explaining}.

To address these security challenges, a wide range of adversarial defense mechanisms have been proposed. These include adversarial training, robust policy optimization, detection algorithms, and input preprocessing techniques \cite{feng2021survey}. However, many of these defenses are developed as standalone solutions and are effective only against specific types of attacks. As a result, they may fail to generalize across different environments or adversarial strategies.

A promising direction is the use of ensemble defense mechanisms, which combine multiple defenses to exploit their complementary strengths \cite{Strauss2017-jo}. Ensemble methods are a well established technique in machine learning for improving generalization and robustness \cite{Strauss2017-jo}. Yet, their application in adversarially robust DRL especially in domains like autonomous driving remains underexplored. By aggregating diverse defenses, ensemble strategies can potentially provide more comprehensive protection against a broader class of attacks, including both white-box and black-box scenarios.

This paper aims to bridge this gap by proposing an ensemble defense framework for DRL, demonstrated in the Highway-env \cite{highway-env} simulation environment. Our approach evaluates the effectiveness of combining multiple defenses to counteract adversarial attacks and highlights how this strategy can significantly enhance the safety and reliability of DRL systems in autonomous driving settings.

\textit{The main contributions of this paper are as follows:}

\begin{itemize}
    \item We propose a novel ensemble defense framework for DRL, combining random noise, autoencoder reconstruction, and Principal Component Analysis (PCA) based filtering to improve robustness against adversarial attacks.

    \item We present a fully inference-time pipeline that operates independently of the policy network, allowing modular deployment without retraining the agent.

    \item Preliminary results show that the ensemble defense outperforms individual defenses on a safety-critical autonomous driving benchmark (Highway-env), both in terms of reward recovery and collision avoidance.
\end{itemize}

To the best of our knowledge, this is the first study to apply an ensemble defense architecture to adversarial robustness in DRL for autonomous driving. By demonstrating the feasibility and advantages of ensemble defenses, this work lays a foundation for more resilient DRL systems, paving the way for safer deployment in real-world applications where reliability is paramount.

\section{Background and Related Work}

In order to discuss further about the ensemble itself, the individual attributes of adversarial attacks and defenses that are available in the literature have to be discussed.

\subsection{Adversarial Attacks}

Adversarial attacks in machine learning are strategies where carefully crafted perturbations often imperceptible to humans are introduced into input data to mislead models into making incorrect predictions. These types of attacks can significantly compromise the reliability of deep learning systems by exposing vulnerabilities in the model’s decision boundaries \cite{RoesslePHP}. In the context of DRL, such attacks typically aim to degrade agent performance by minimizing the cumulative reward, all while remaining undetected by standard observation mechanisms \cite{Ali2023-oc}.

Recent works have demonstrated the efficacy of both white-box and black-box attack methods against DRL agents. For instance, \cite{135} proposed the use of high-confidence adversarial examples created via environmental obstacles, achieving a remarkable 99.9\% Attack Success Rate (ASR) in a pathfinding task under white-box conditions using the Asynchronous Advantage Actor-Critic (A3C) algorithm.

Among gradient-based approaches, the Fast Gradient Sign Method (FGSM) \cite{140} remains widely studied due to its simplicity and effectiveness across time steps particularly in Atari game environments though it is more prone to detection. More advanced variants, such as Policy Induction Attacks \cite{142}, combine FGSM with Jacobian-based Saliency Map Attacks to improve transferability across DQN variants, while also outperforming random noise at lower perturbation rates \cite{143}. 

Physical systems such as autonomous robots have also been targeted, where adversarial perturbations to sensory inputs can severely compromise navigation \cite{144}. Universal mask attacks like CopyCAT \cite{145} demonstrate the feasibility of both targeted and untargeted disruptions using deep Q-learning frameworks. Likewise, the ACADIA framework \cite{146} introduces momentum-based perturbations with adaptive estimation techniques, pushing the limits of white-box and black-box effectiveness.

Strategic attacks in multi-agent environments \cite{150}, timing-based attacks \cite{141}, and transformation-based methods like Adversarial Transformation Networks \cite{154} further diversify the threat landscape. Notably, backdoor attacks such as TrojDRL \cite{155} embed hidden triggers during training, and novel techniques like Critical Point \cite{147}, Action Poisoning \cite{156}, and Tentative Frame Attacks \cite{157} aim to optimize disruption using minimal yet impactful perturbations.

\subsection{Adversarial Defenses}

In response to the growing threat of adversarial attacks, a wide array of defense mechanisms have emerged to safeguard DRL agents. These defenses generally aim to either detect adversarial inputs or enhance the agent’s resilience during training and deployment.

One of the most foundational strategies is Adversarial Training (AT), where agents are explicitly trained on perturbed inputs to improve robustness. Variants of AT have demonstrated strong performance across settings, especially when combined with advanced perturbation techniques \cite{135,151,162,163}. Robust Reinforcement Learning, through formulations like noisy action robust MDPs \cite{160}, provides a principled framework for handling uncertainty in both adversarial and non-adversarial environments.

Detection-based methods, such as those using Principal Component Analysis (PCA) \cite{164,165}, are useful for identifying potential attacks without necessarily altering the policy. Meanwhile, regularization techniques \cite{167,168} and imitation learning frameworks \cite{166} stabilize learning and reduce the effect of adversarial drift by guiding the policy towards safe behavior.

More advanced defenses incorporate generative or predictive elements. RADIAL \cite{171,172,173} and RL-VAEGAN \cite{171} leverage generative models to distinguish between clean and adversarial states. Frame prediction techniques have also shown promise, detecting adversarial inputs with up to 100\% accuracy in some benchmarks \cite{164}.

Collectively, these defense strategies reflect a broader shift toward proactive robustness in DRL, integrating detection, adaptation, and protection mechanisms to ensure reliable performance even in adversarial environments.

\subsection{Ensemble of Defenses}

Ensemble defenses in deep neural networks (DNN) involve training multiple base classifiers collaboratively to enhance robustness against adversarial attacks. In DRL, the concept of an ensemble of defenses against adversarial attacks is different from the classical ensemble approaches used in DNN for classification. \cite{Haydari2021-az} discusses ensemble-based adversarial detection techniques for Traffic Controllers. To the best of our knowledge, no prior work on the topic of the ensemble of defenses for autonomous driving in DRL exists in the literature.

\section{Experimental Setup}

The Highway-env \cite{highway-env} is a collection of scenarios to train and test DRL agents in autonomous driving like \emph{Merge}, \emph{Intersection}, and \emph{Roundabout}.

In this paper, we choose the Highway scenario to experiment with the ensemble of defenses idea in autonomous driving scenarios. In the Highway scenario figure \ref{fig:ensemble_defense}, the state space is continuous, and we choose discrete meta-actions, namely $a=\{0:\text{Lane\_left}, 1:\text{Idle}, 2:\text{Lane\_right}, 3:\text{Faster}, 4:\text{Slower}\}$ \cite{Karpenahalli_Ramakrishna2025-nv}.

\begin{figure}[ht]
    \centering
    \includegraphics[width=1.0\linewidth]{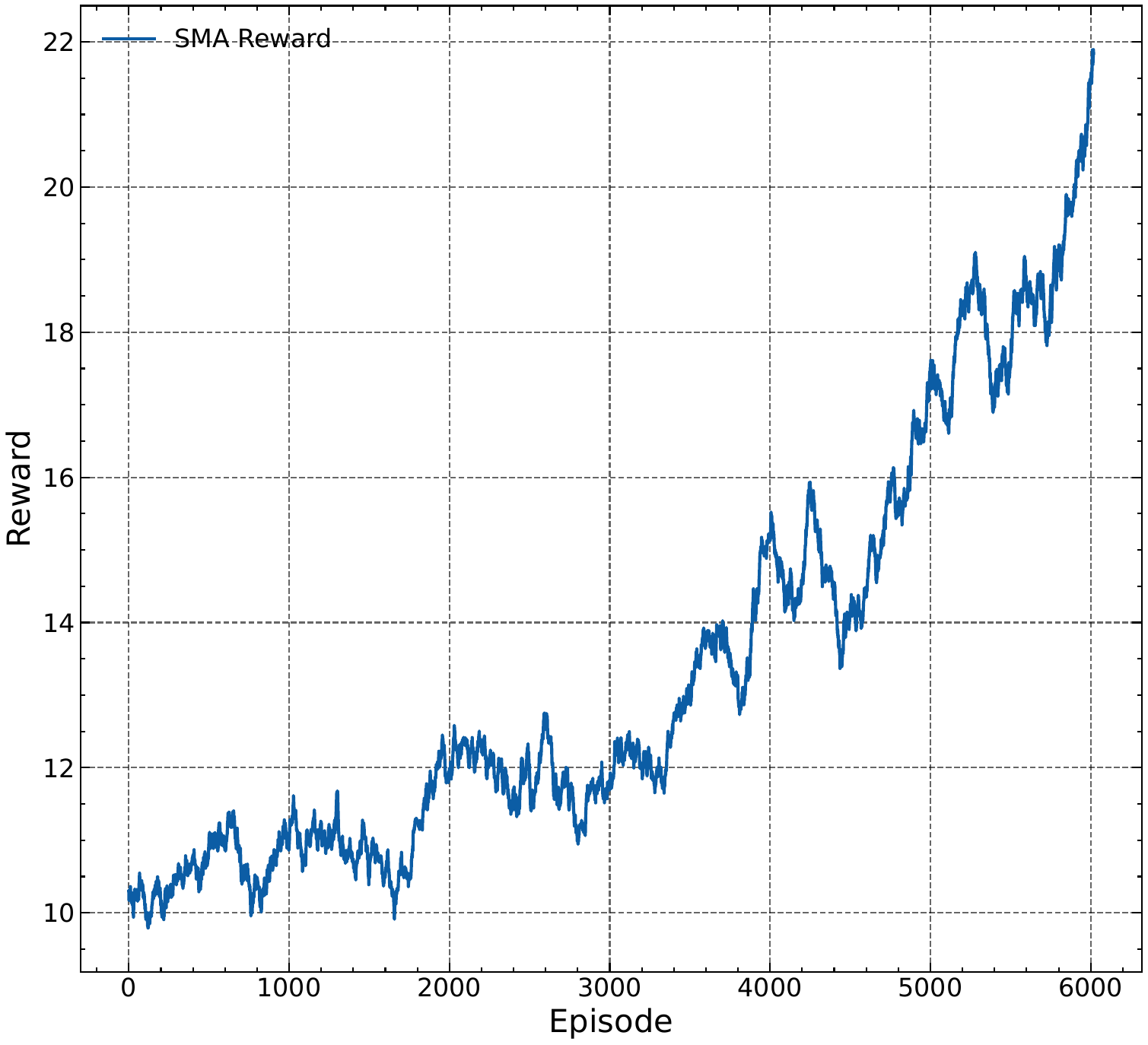}
    \caption{Simple Moving Average (SMA) of rewards during DQN training across 6000 episodes. The agent learns to maximize long-term return, converging steadily. This trained policy is used for adversarial robustness evaluation.}
    \label{fig:dqn_sma_training_curve}
\end{figure}

We begin our pipeline by training a vanilla Deep Q-Network (DQN) agent on the clean, unperturbed Highway scenario. The DQN agent is implemented using Stable-Baselines3 \cite{Raffin2021-mv}, and training is conducted for 6000 episodes. The purpose of this model is to serve as the "nominal expert" against which adversarial perturbations and defense mechanisms are evaluated.

After the Vanilla DQN is trained, the environment is subjected to adversarial perturbations based on FGSM during inference, and multiple defenses are evaluated. The data collected during clean policy rollouts is used to train the autoencoder and PCA modules. Specifically, the autoencoder is trained in an offline, unsupervised fashion using reconstruction Mean Squared Error loss over 5000 sampled observations.

All evaluations and visualizations are performed over 100 episodes per scenario. For consistency, the random seeds are fixed for sampling, environment resets, and action selections.
\subsection{Adversarial Attacks}

To evaluate the robustness of our DRL agents, we implement adversarial attacks focused on state perturbations. In particular, we employ FGSM, a gradient-based method known for its effectiveness in fooling neural network-based agents with minimal perturbations.

\subsubsection{Fast Gradient Sign Method (FGSM)}
FGSM introduces noise to the input state in the direction of the gradient of the loss function with respect to the input. Given a state $s \in \mathbb{R}^n$, the perturbed adversarial state $s_{\text{adv}}$ is computed as:
\begin{equation}
    s_{\text{adv}} = s + \epsilon \cdot \text{sign}\left(\nabla_s J(\pi_\theta, s, a)\right),
\end{equation}
where $\epsilon$ is the attack magnitude, $J$ is the loss function, $\pi_\theta$ the policy parameterized by $\theta$, and $a$ is the selected action. FGSM serves as a standard benchmark for testing robustness in adversarially perturbed DRL.


\subsection{Defensive Mechanisms}

We evaluate the effectiveness of multiple defense mechanisms designed to mitigate adversarial perturbations. Each defense operates on the input state and attempts to recover clean or meaningful information before it is passed to the agent.

\subsubsection{Random Noise Defense}
This defense introduces small, uniformly distributed noise to counterbalance adversarial perturbations:
\begin{equation}
    s_{\text{def}} = \text{clip}(s + \mathcal{U}(-\eta, \eta), 0, 1),
\end{equation}
where $\eta$ is a small positive constant.

\subsubsection{Autoencoder Defense}
A shallow autoencoder is trained on clean (nominal) observations collected from the environment. During inference, the autoencoder reconstructs the clean state from a potentially perturbed input, acting as a denoising filter. The autoencoder consists of a fully connected encoder-decoder architecture, taking a flattened $5 \times 5$ (25-dimensional) input. The encoder maps the input through layers of size 128 and 64, while the decoder reconstructs it back via layers of size 128 and 25, using ReLU activations in the hidden layers and mean squared error loss during training.

\subsubsection{PCA-Based Defense}
PCA is used to project the input state onto a lower-dimensional subspace and reconstruct it. This projection inherently suppresses noise by emphasizing dominant features, which aids in mitigating adversarial signals.

\subsubsection{Ensemble Defense}
We propose a novel ensemble defense framework that integrates the above methods. At runtime, the ensemble takes the mean of the corrected states from each individual defense:
\begin{equation}
    s_{\text{ensemble}} = \frac{1}{3} (s_{\text{random}} + s_{\text{autoencoder}} + s_{\text{pca}}).
\end{equation}
This aggregated state balances the strengths and weaknesses of individual defenses.

\begin{algorithm}[ht]
\caption{\textbf{Ensemble Defense Pipeline at Inference}}
\label{alg:ensemble-defense}
\begin{algorithmic}[1]
\Require Trained DQN policy $\pi_\theta$, environment $\mathcal{E}$, number of episodes $N$, FGSM Noise $\epsilon$, Cumulative Rewards $R$, 
\Ensure Episode-wise cumulative reward and collision rate
\State Initialize defense modules: \texttt{RandomNoise}, \texttt{Autoencoder}, \texttt{PCA}
\For{$e = 1$ to $N$}
    \State $s \gets \mathcal{E}.reset()$, $done \gets \text{False}$, $R \gets 0$, $collided \gets \text{False}$
    \While{not $done$}
        \State $\tilde{s} \gets s + \epsilon \cdot \text{sign}(\nabla_s J(\pi_\theta, s))$ \Comment{Apply FGSM attack}
        \State $s_r \gets \texttt{RandomNoise}(\tilde{s})$
        \State $s_{ae} \gets \texttt{Autoencoder}(\tilde{s})$
        \State $s_{pca} \gets \texttt{PCA}(\tilde{s})$
        \State $\hat{s} \gets \frac{1}{3}(s_r + s_{ae} + s_{pca})$ \Comment{Ensemble fusion}
        \State $a \gets \pi_\theta(\hat{s})$
        \State $s', r, done, info \gets \mathcal{E}.step(a)$
        \State $R \gets R + r$
        \If{collision occurred in $info$}
            \State $collided \gets \text{True}$
        \EndIf
        \State $s \gets s'$
    \EndWhile
    \State Log $(R, collided)$
\EndFor
\State \Return All $(R, collided)$ pairs
\end{algorithmic}
\end{algorithm}

\noindent
Algorithm~\ref{alg:ensemble-defense} outlines the full inference-time defense pipeline used to evaluate the robustness of a fixed DQN policy under adversarial attacks. At each timestep, the incoming perturbed observation $s$ is passed through three parallel defenses. Their outputs are fused via averaging to form a robust estimate of the clean input, which is then passed to the trained policy for action selection. This procedure is repeated for $N$ episodes, logging cumulative reward and safety (collision) statistics. Importantly, the agent's policy remains untouched, highlighting the modular, plug-and-play nature of the ensemble defense.


\section{Results and Discussions}

The preliminary evaluation was conducted over 100 episodes per scenario, comparing both performance (reward) and safety metrics (collision rate). Each defense strategy is assessed under FGSM perturbation and compared against the clean policy and attack-only baseline. Results were logged and aggregated into statistical summaries, smoothed reward trajectories, and distribution visualizations.

 \begin{figure}[ht]
    \centering
    \includegraphics[width=\linewidth]{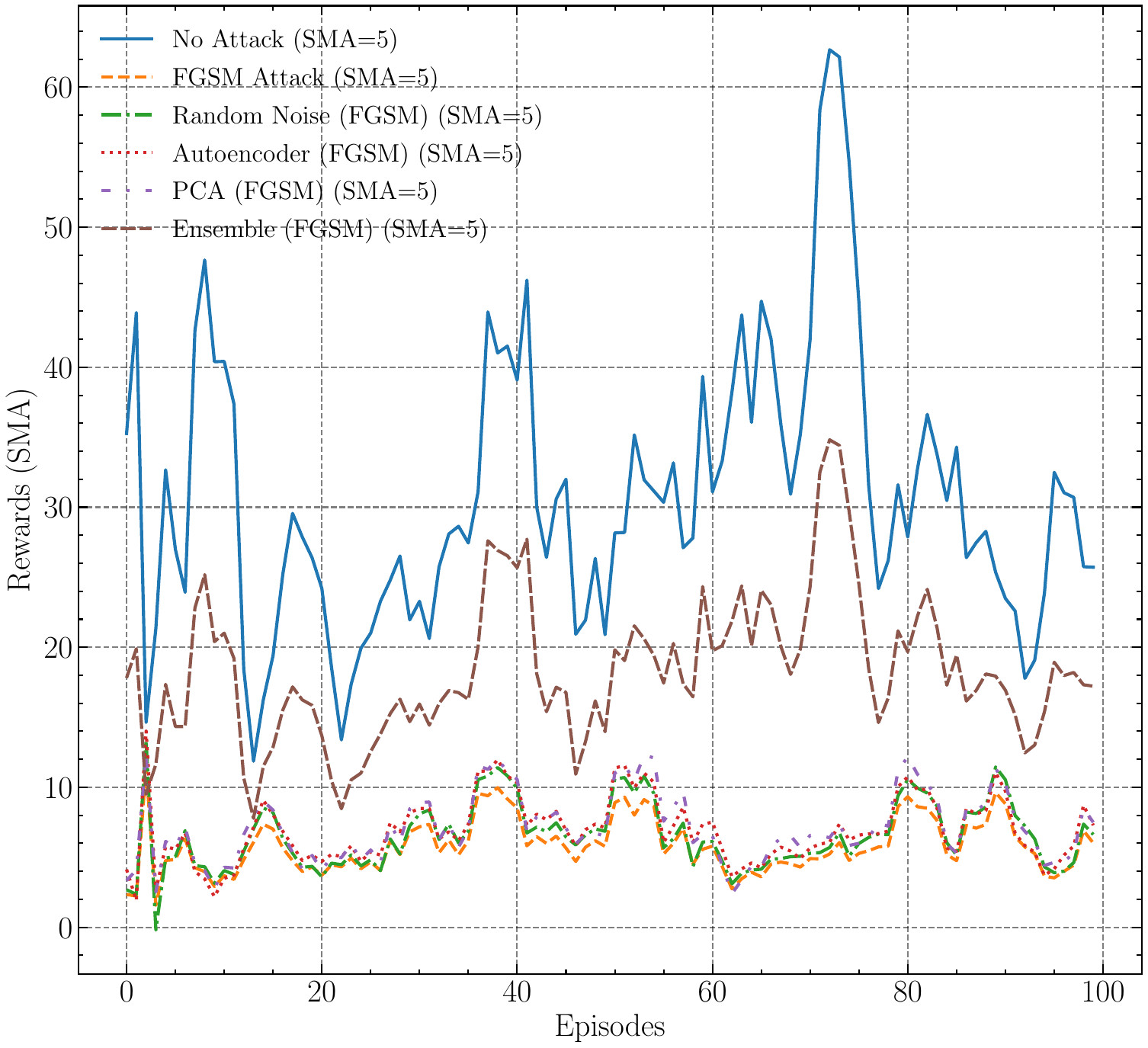}
    \caption{SMA of Rewards across 100 episodes under various adversarial scenarios. The Ensemble defense demonstrates a significant improvement in robustness compared to individual defenses.}
    \label{fig:sma_rewards_plot}
\end{figure}

\begin{figure}[ht]
    \centering
    \includegraphics[width=\linewidth]{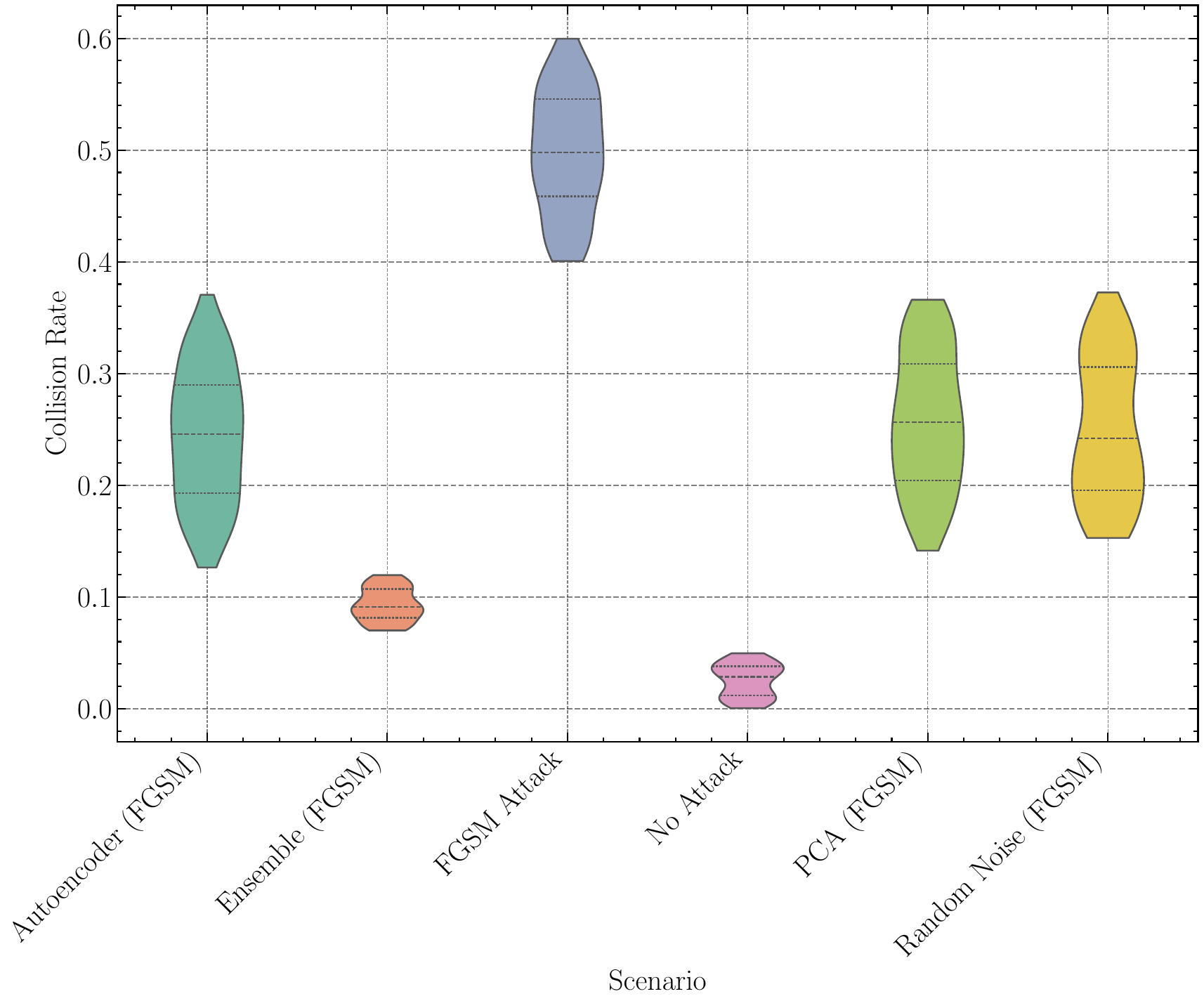}
    \caption{Violin plot illustrating the distribution of collision rates across different scenarios. Ensemble defense not only improves reward but also significantly reduces variance and frequency of collisions.}
    \label{fig:collision_violin_plot}
\end{figure}

\begin{table}[ht]
\caption{Statistical Summary of Results - Highway scenario}
\label{tab:summary_stats}
\centering
\resizebox{\linewidth}{!}{%
\begin{tabular}{|l|c|c|c|c|}
\hline
\textbf{Scenario} & \textbf{Mean Reward} & \textbf{Std Reward} & \textbf{Mean Collision Rate} & \textbf{Std Collision Rate} \\
\hline
No Attack (Baseline) & 30.63 & 20.99 & 0.03 & 0.01\\
\hline
FGSM Attack & 5.87 & 4.13 & 0.50 & 0.05 \\
Random Noise (FGSM) & 6.57 & 4.77 & 0.25 & 0.06 \\
Autoencoder (FGSM) & 6.94 & 4.86 & 0.24 & 0.06 \\
PCA (FGSM) & 7.15 & 5.34 & 0.26 & 0.06 \\
\textbf{Ensemble (FGSM)} & \textbf{18.38} & \textbf{11.26} & \textbf{0.09} & \textbf{0.01} \\
\hline
\end{tabular}%
}
\end{table}

From Table~\ref{tab:summary_stats}, we observe that adversarial perturbations via FGSM dramatically reduce the policy's performance. Specifically, the mean reward under FGSM attack drops by over 80\% compared to the clean baseline, while the mean collision rate surges from 0.03 to 0.50. Each standalone defense whether Random Noise, Autoencoder, or PCA provides only marginal mitigation, offering moderate improvements in rewards and slight reductions in collision rates. However, their high standard deviations indicate instability and inconsistent protection.

In contrast, the ensemble approach achieves a mean reward of 18.38, recovering more than 60\% of the original performance while keeping the collision rate down to just 0.09. Notably, this recovery is achieved without retraining the policy or explicitly learning on adversarial examples.

The SMA reward plot in Figure~\ref{fig:sma_rewards_plot} clearly illustrates the performance trend: ensemble defense curves are distinctly higher and more stable across episodes compared to other defenses. Similarly, the violin plot in Figure~\ref{fig:collision_violin_plot} shows a condensed, narrow distribution of collisions in the ensemble case (orange), indicating consistent safety.

Importantly, while the ensemble significantly improves robustness under attack, its reward curve does not fully match the no-attack baseline. This performance gap indicates that the adversarial perturbations remain active during inference, and although the ensemble filters help mitigate their effect, the agent is still operating under continuous threat. Therefore, complete recovery to baseline levels is inherently limited under persistent adversarial conditions. The hypothesis is that each defense mechanism in the ensemble contributes a unique capability:

\begin{itemize}
    \item Random Noise: Helps in breaking small adversarial patterns but introduces stochasticity that may reduce precision.
    \item Autoencoder: Trained to reconstruct clean features, helping recover semantic structures lost in adversarial perturbation.
    \item PCA: Filters input to retain dominant components and project noise into a lower-dimensional, less harmful space.
\end{itemize}

While these methods show individual merit, none consistently outperform the ensemble, which balances the signal recovery strengths across methods. From a robustness perspective, the ensemble can be seen as a filtering consensus mechanism: each component processes the input differently, and their aggregation ensures that even if one fails on a given input, the others can compensate. This redundancy is particularly effective in environments where the nature of perturbation may vary or be unknown. Additionally, the ensemble design is modular and independent of the agent architecture. It is applied solely at the observation level, making it deployable without retraining or modifying the base policy.

\begin{table}[ht]
\caption{Statistical Summary of Results - Merge scenario}
\label{tab:summary_merge_stats}
\centering
\resizebox{\linewidth}{!}{%
\begin{tabular}{|l|c|c|c|c|}
\hline
\textbf{Scenario} & \textbf{Mean Reward} & \textbf{Std Reward} & \textbf{Mean Collision Rate} & \textbf{Std Collision Rate} \\
\hline
No Attack (Baseline) & 11.93 & 0.50 & 0.01 & 0.01 \\
\hline
FGSM Attack & 2.90 & 0.92 & 0.69 & 0.10 \\
Random Noise (FGSM) & 7.85 & 1.19 & 0.39 & 0.08 \\
Autoencoder (FGSM) & 7.83 & 1.12 & 0.39 & 0.08 \\
PCA (FGSM) & 8.07 & 1.19 & 0.39 & 0.08 \\
\textbf{Ensemble (FGSM)} & \textbf{11.49} & \textbf{0.28} & \textbf{0.02} & \textbf{0.01} \\
\hline
\end{tabular}%
}
\end{table}

To further validate the generalizability of our findings, we extended the evaluation to the merge scenario from Highway-Env environment \cite{highway-env}. Compared to highway scenario, merge scenario presents a simpler decision space and fewer dynamic interactions, allowing clearer observation of defense behavior under perturbation.

As shown in Table~\ref{tab:summary_merge_stats}, the results are even more promising. The baseline (No Attack) policy achieves a stable mean reward of 11.93 with minimal variance and a very low collision rate of 0.01. Under FGSM attack, the reward plummets to 2.90 and the collision rate spikes to 0.69, highlighting the attack's disruptive nature.

Individual defenses again provide only partial mitigation, with mean rewards ranging between 7.83 and 8.07, and collision rates near 0.39. However, the ensemble approach recovers a substantial portion of the baseline performance, reaching a mean reward of 11.49 and reducing the collision rate to 0.02, almost matching the no-attack case.

Interestingly, both the mean and standard deviations for ensemble performance are tighter in merge scenario compared to highway scenario. This stability suggests that the ensemble defense benefits even more from the relatively deterministic structure of the merging task. The simpler dynamics allow the ensemble to consistently recover from perturbations, reinforcing its role as a practical and robust defense mechanism across varying task complexities.


\section{Conclusion}

Both the results demonstrate that the ensemble defense improves agent robustness under adversarial perturbations in autonomous driving, recovering a significant portion of the original performance. Notably, the mean reward achieved under ensemble defense does not fully match that of the clean (no-attack) baseline. This is attributed to the fact that the adversarial attack remains active during inference, and the agent must operate under continuous perturbation. The observed performance, therefore, reflects the effectiveness of the defense in mitigating the influence of adversarial noise rather than eliminating it. The agent’s ability to maintain stable behavior and reduced collision rates under such conditions highlights the practical utility of ensemble defenses in real-world, safety-critical scenarios where retraining or adversarial training may not be feasible.

Our current setup focuses solely on FGSM state perturbations. Future experiments will extend this to include PGD, CW attacks, and black-box transfer settings. We also aim to:
\begin{itemize}

    \item Introduce adaptive ensemble weighting strategies conditioned on input variance.
    \item Explore combining ensemble defenses with robust training techniques.
    \item Extend the experiment to multiple RL algorithms and do more ablation study.
    \item Evaluate performance on multi-agent environments and real-world simulations.
\end{itemize}

These enhancements aim to validate the generalizability, scalability, and efficiency of ensemble defenses across broader DRL deployments.



\bibliographystyle{IEEEtran}
\bibliography{mybibfile}

\begin{thebibliography}{10}
\providecommand{\url}[1]{#1}
\csname url@samestyle\endcsname
\providecommand{\newblock}{\relax}
\providecommand{\bibinfo}[2]{#2}
\providecommand{\BIBentrySTDinterwordspacing}{\spaceskip=0pt\relax}
\providecommand{\BIBentryALTinterwordstretchfactor}{4}
\providecommand{\BIBentryALTinterwordspacing}{\spaceskip=\fontdimen2\font plus
\BIBentryALTinterwordstretchfactor\fontdimen3\font minus \fontdimen4\font\relax}
\providecommand{\BIBforeignlanguage}[2]{{%
\expandafter\ifx\csname l@#1\endcsname\relax
\typeout{** WARNING: IEEEtran.bst: No hyphenation pattern has been}%
\typeout{** loaded for the language `#1'. Using the pattern for}%
\typeout{** the default language instead.}%
\else
\language=\csname l@#1\endcsname
\fi
#2}}
\providecommand{\BIBdecl}{\relax}
\BIBdecl

\bibitem{Ilahi2022-bq}
I.~Ilahi, M.~Usama, J.~Qadir, M.~U. Janjua, A.~Al-Fuqaha, D.~T. Hoang, and D.~Niyato, ``Challenges and countermeasures for adversarial attacks on deep reinforcement learning,'' \emph{IEEE Transactions on Artificial Intelligence}, vol.~3, no.~2, pp. 90--109, Apr. 2022.

\bibitem{feng2021survey}
C.~Feng, L.~Xu, Y.~Wang, and J.~Wu, ``A survey on adversarial attacks and defenses in reinforcement learning,'' \emph{Artificial Intelligence Review}, vol.~55, no.~4, pp. 2689--2711, 2021.

\bibitem{nguyen2019deep}
T.-T. Nguyen, T.~L. Nguyen, and S.~Nahavandi, ``Deep reinforcement learning: An overview,'' \emph{arXiv preprint arXiv:1906.07006}, 2019.

\bibitem{lin2020review}
J.~Lin, S.~H.~M. Ahmad, A.~Khattak, and J.~Wang, ``A review of applications of artificial intelligence and blockchain in the energy sector,'' \emph{Energies}, vol.~13, no.~14, p. 3652, 2020.

\bibitem{huang2017adversarial}
S.~Huang, N.~Papernot, I.~Goodfellow, Y.~Duan, and P.~Abbeel, ``Adversarial attacks on neural network policies,'' in \emph{Workshop on artificial intelligence safety 2017 (AISafety 2017)}.\hskip 1em plus 0.5em minus 0.4em\relax PMLR, 2017, pp. 1--8.

\bibitem{behzadan2017whatever}
V.~Behzadan and A.~Munir, ``Whatever does not kill deep reinforcement learning, makes it stronger,'' \emph{arXiv preprint arXiv:1711.09344}, 2017.

\bibitem{evtimov2017robust}
K.~Eykholt \emph{et~al.}, ``Robust physical-world attacks on deep learning visual classification,'' in \emph{Proceedings of the IEEE Conference on Computer Vision and Pattern Recognition (CVPR)}, 2018.

\bibitem{goodfellow2014explaining}
I.~J. Goodfellow, J.~Shlens, and C.~Szegedy, ``Explaining and harnessing adversarial examples,'' in \emph{International Conference on Learning Representations (ICLR)}, 2015.

\bibitem{Strauss2017-jo}
T.~Strauss, M.~Hanselmann, A.~Junginger, and H.~Ulmer, ``Ensemble methods as a defense to adversarial perturbations against deep neural networks,'' \emph{arXiv preprint arXiv:1709.03423}, 2017.

\bibitem{highway-env}
E.~Leurent, ``An environment for autonomous driving decision-making,'' \url{https://github.com/eleurent/highway-env}, 2018.

\bibitem{RoesslePHP}
D.~R{\"o}{\ss}le, D.~Cremers, and T.~Sch{\"o}n, ``Perceiver hopfield pooling for dynamic multi-modal and multi-instance fusion,'' in \emph{Artificial Neural Networks and Machine Learning -- ICANN 2022}.\hskip 1em plus 0.5em minus 0.4em\relax Cham: Springer International Publishing, 2022, pp. 599--610.

\bibitem{Ali2023-oc}
H.~Ali, D.~Chen, M.~Harrington, N.~Salazar, M.~Al~Ameedi, A.~Khan, A.~R. Butt, and J.-H. Cho, ``A survey on attacks and their countermeasures in deep learning: Applications in deep neural networks, federated, transfer, and deep reinforcement learning,'' \emph{IEEE Access}, vol.~11, pp. 120\,095--120\,130, 2023.

\bibitem{135}
T.~Chen, W.~Niu, Y.~Xiang, X.~Bai, J.~Liu, Z.~Han, and G.~Li, ``Gradient band-based adversarial training for generalized attack immunity of a3c path finding,'' \emph{arXiv preprint arXiv:1807.06752}, 2018.

\bibitem{140}
S.~Huang, N.~Papernot, I.~Goodfellow, Y.~Duan, and P.~Abbeel, ``Adversarial attacks on neural network policies,'' in \emph{Proc. ICLR}, 2017, pp. 1--10.

\bibitem{142}
V.~Behzadan and A.~Munir, ``Vulnerability of deep reinforcement learning to policy induction attacks,'' in \emph{Proc. Int. Conf. Mach. Learn. Data Mining Pattern Recognit.}, 2017, pp. 262--275.

\bibitem{143}
J.~Kos and D.~Song, ``Delving into adversarial attacks on deep policies,'' in \emph{Proc. 34th Int. Conf. Mach. Learn.}, vol.~70, 2017, pp. 1944--1953.

\bibitem{144}
G.~Clark, M.~Doran, and W.~Glisson, ``A malicious attack on the machine learning policy of a robotic system,'' in \emph{Proc. 17th IEEE Int. Conf. Trust, Secur. Privacy Comput. Commun./12th IEEE Int. Conf. Big Data Sci. Eng.}, 2018, pp. 516--521.

\bibitem{145}
L.~Hussenot, M.~Geist, and O.~Pietquin, ``Copycat: Taking control of neural policies with constant attacks,'' in \emph{Int. Found. Auton. Agents Multiagent Syst.}, Auckland, New Zealand, 2019.

\bibitem{146}
H.~Ali, M.~A. Ameedi, A.~Swami, R.~Ning, J.~Li, H.~Wu, and J.-H. Cho, ``Acadia: Efficient and robust adversarial attacks against deep reinforcement learning,'' in \emph{Proc. IEEE Conf. Commun. Netw. Secur. (CNS)}, 2022, pp. 1--9.

\bibitem{150}
M.~Figura, K.~C. Kosaraju, and V.~Gupta, ``Adversarial attacks in consensus-based multi-agent reinforcement learning,'' in \emph{Proc. Amer. Control Conf. (ACC)}, 2021, pp. 3050--3055.

\bibitem{141}
Y.-C. Lin, Z.-W. Hong, Y.-H. Liao, M.-L. Shih, M.-Y. Liu, and M.~Sun, ``Tactics of adversarial attack on deep reinforcement learning agents,'' in \emph{Proc. 26th Int. Joint Conf. Artif. Intell.}, 2017, pp. 3756--3762.

\bibitem{154}
E.~Tretschk, S.~J. Oh, and M.~Fritz, ``Sequential attacks on agents for long-term adversarial goals,'' in \emph{Proc. ACM Comput. Sci. Cars Symp.}, 2018, pp. 1--9.

\bibitem{155}
K.~Panagiota, W.~Kacper, S.~Jha, and L.~Wenchao, ``Trojdrl: Trojan attacks on deep reinforcement learning agents,'' in \emph{Proc. 57th ACM/IEEE Design Automat. Conf. (DAC)}, 2020, pp. 1--17.

\bibitem{147}
J.~Sun, T.~Zhang, X.~Xie, L.~Ma, Y.~Zheng, K.~Chen, and Y.~Liu, ``Stealthy and efficient adversarial attacks against deep reinforcement learning,'' in \emph{Proc. AAAI Conf. Artif. Intell.}, vol.~34, 2020, pp. 5883--5891.

\bibitem{156}
G.~Liu and L.~Lai, ``Provably efficient black-box action poisoning attacks against reinforcement learning,'' in \emph{Proc. Adv. Neural Inf. Process. Syst.}, vol.~34, 2021, pp. 12\,400--12\,410.

\bibitem{157}
Q.~Ye, X.~Zhou, C.~Ying, and J.~Zhu, ``Strategically-timed state-observation attacks on deep reinforcement learning agents,'' in \emph{Proc. Int. Conf. Mach. Learn. (ICML)}, 2021, pp. 1--10.

\bibitem{151}
A.~Pattanaik, Z.~Tang, S.~Liu, G.~Bommannan, and G.~Chowdhary, ``Robust deep reinforcement learning with adversarial attacks,'' in \emph{Proc. 17th Int. Conf. Auto. Agents MultiAgent Syst.}, 2018, pp. 2040--2042.

\bibitem{162}
Q.~Shen, Y.~Li, H.~Jiang, Z.~Wang, and T.~Zhao, ``Deep reinforcement learning with robust and smooth policy,'' in \emph{Proc. 37th Int. Conf. Mach. Learn.}, 2020, pp. 8707--8718.

\bibitem{163}
H.~Zhang, H.~Chen, D.~S. Boning, and C.-J. Hsieh, ``Robust reinforcement learning on state observations with learned optimal adversary,'' in \emph{Proc. Int. Conf. Learn. Represent.}, 2021, pp. 1--16.

\bibitem{160}
C.~Tessler, Y.~Efroni, and S.~Mannor, ``Action robust reinforcement learning and applications in continuous control,'' in \emph{Proc. Int. Conf. Mach. Learn.}, 2019, pp. 6215--6224.

\bibitem{164}
Y.-C. Lin, M.-Y. Liu, M.~Sun, and J.-B. Huang, ``Detecting adversarial attacks on neural network policies with visual foresight,'' \emph{arXiv preprint arXiv:1710.00814}, 2017.

\bibitem{165}
Y.~Xiang, W.~Niu, J.~Liu, T.~Chen, and Z.~Han, ``A pca-based model to predict adversarial examples on q-learning of path finding,'' in \emph{Proc. IEEE 3rd Int. Conf. Data Sci. Cyberspace (DSC)}, 2018, pp. 773--780.

\bibitem{167}
A.~Russo and A.~Proutiere, ``Optimal attacks on reinforcement learning policies,'' \emph{arXiv preprint arXiv:1907.13548}, 2019.

\bibitem{168}
H.~Zhang, H.~Chen, C.~Xiao, B.~Li, M.~Liu, D.~Boning, and C.-J. Hsieh, ``Robust deep reinforcement learning against adversarial perturbations on state observations,'' in \emph{Proc. Adv. Neural Inf. Process. Syst. (NIPS)}, vol.~33, 2020, pp. 21\,024--21\,037.

\bibitem{166}
M.~Fischer, M.~Mirman, S.~Stalder, and M.~Vechev, ``Online robustness training for deep reinforcement learning,'' \emph{arXiv preprint arXiv:1911.00887}, 2019.

\bibitem{171}
Y.~Hu and S.~Sun, ``Rl-vaegan: Adversarial defense for reinforcement learning agents via style transfer,'' \emph{Knowl.-Based Syst.}, vol. 221, p. 106967, 2021.

\bibitem{172}
B.~Lütjens, M.~Everett, and J.~P. How, ``Certified adversarial robustness for deep reinforcement learning,'' in \emph{Proc. Conf. Robot Learn.}, 2020, pp. 1328--1337.

\bibitem{173}
T.~Oikarinen, W.~Zhang, A.~Megretski, L.~Daniel, and T.-W. Weng, ``Robust deep reinforcement learning through adversarial loss,'' in \emph{Proc. Adv. Neural Inf. Process. Syst.}, vol.~34, 2021, pp. 26\,156--26\,167.

\bibitem{Haydari2021-az}
A.~Haydari, M.~Zhang, and C.-N. Chuah, ``Adversarial attacks and defense in deep reinforcement learning ({DRL})-based traffic signal controllers,'' \emph{IEEE Open J. Intell. Transp. Syst.}, vol.~2, pp. 402--416, 2021.

\bibitem{Karpenahalli_Ramakrishna2025-nv}
C.~Karpenahalli~Ramakrishna, A.~Mohan, Z.~Zeinaly, and L.~Belzner, ``The evolution of criticality in deep reinforcement learning,'' in \emph{Proceedings of the 17th International Conference on Agents and Artificial Intelligence}.\hskip 1em plus 0.5em minus 0.4em\relax SCITEPRESS - Science and Technology Publications, 2025, pp. 217--224.

\bibitem{Raffin2021-mv}
A.~Raffin, A.~Hill, A.~Gleave, A.~Kanervisto, M.~Ernestus, and N.~Dormann, ``Stable-{Baselines3}: Reliable reinforcement learning implementations,'' \emph{J. Mach. Learn. Res.}, vol.~22, no. 268, pp. 1--8, 2021.

\end{thebibliography}


\end{document}